\documentclass{article}
\usepackage{spconf, amsmath, epsfig}

\title{Object detection in satellite imagery using 2-Step Convolutional Neural Networks}
\name{Hiroki Miyamoto$^*$$^1$\thanks{$^*$miyamoto-hrk-tomeken@aist.go.jp}, Kazuki Uehara$^1$, Masahiro Murakawa$^1$}
\names{Hidenori Sakanashi$^1$, Hirokazu Nosato$^1$, Toru Kouyama$^1$, Ryosuke Nakamura$^1$} 
\address{$^1$National Institute of Advanced Industrial Science and Technology, Tokyo, Japan}

\begin{document}
%\ninept
%
\maketitle
\begin{abstract}
This paper presents an efficient object detection method from satellite imagery. Among a number of machine learning algorithms, we proposed a combination of two convolutional neural networks (CNN)  aimed at high precision and high recall, respectively. We validated our models using golf courses as target objects. The proposed deep learning method demonstrated higher accuracy than previous object identification methods.

\end{abstract}
\section{Introduction}
\label{sec:intro}Earth observation satellites have been monitoring changes on the Earth's surface over a long period of time. High resolution satellite imagery can detect small objects such as ships, cars, aircraft,  and individual houses; whereas, medium resolution satellite imagery can detect relatively larger objects, such as ports, roads, airports and large buildings [1][6]. Total data amount, however, would be too huge to be inspected by human eyes.  Therefore, we need an efficient algorithm for automatic object detection on satellite imagery. Previous works employed higher-order local auto correlation [1], random forests [2], and deep learning [3][4][5][6]. Among them, deep learning [7][8] showed higher accuracy in object detection of images than other machine learning methods.

As an example of target object, we selected golf courses because they exist everywhere in the world, are typically of a recognizable size and shape with 30 meter resolution of Landsat 8 imagery. According to the R \& A report [9], in 2016 there were 33161 golf courses, which provide more than enough data for training and detection purpose. Once we establish an accurate algorithm, we can continuously monitor the new construction and disappearance of all the golf courses on the Earth.

Among the general object detection framework, Faster R-CNN [8] is the state-of-the-art method. This method consists of a region proposal network for predicting candidate regions and region classification network for classifying object proposal. It is an end-to-end detector that outputs the location and category of object simultaneously. Similarly, we propose a model called “high recall network” specifically for detection of candidate golf course regions and a model called “high precision network” for further confirmation. We then compared the proposed method with other existing methods [1][6].

\section{METHOD}
\label{sec:format}
The framework of our object detection method, which involved a two-step process, is illustrated in Fig. 1. The first step employs the high recall network (HRN) model to find candidate regions as much as possible. The second step uses the high precision network (HPN) model for binary classification (golf course or not) of the HRN output. Each step is customized for the purpose as described in 2.1 and 2.2.

\begin{figure*}[t]
  \begin{center}
\vspace{-4cm}
  \includegraphics[bb=0 0 1913 682, scale=0.43]{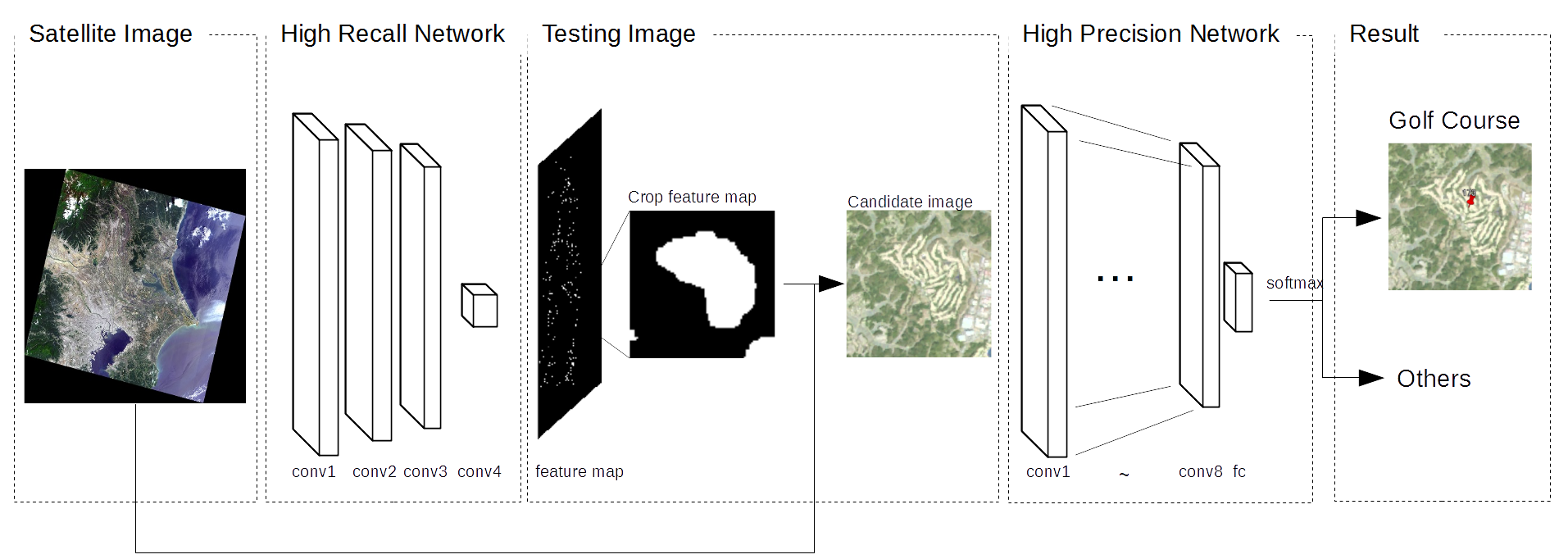}
  \caption{The flowchart of our 2-step CNN model}
  \label{}
  \end{center}
\end{figure*}

\subsection{High Recall Network (HRN) }
The training and validation data set for HRN is derived from  117 clear Landsat 8 scenes taken between 2013 and 2015 all over Japan except five areas reserved for testing (see 3.1 for more detail). Each area is observed 3 to 4 times at different time to account for seasonal variation. Moreover, we have prepared Ground Truth (GT) polygons which outline all the golf courses in Japan. The Landsat scenes were gridded into tiles with 16$\times$16 pixels. A tile is classified as positive if the coverage of GT polygons is larger than 20\% of the total area; tiles with no overlap with golf courses were classified as negative; and the remainder which fell between 0\% $\sim$ 20\% coverage were classified as neither positive nor negative (Fig. 2). 

Ishii et al. [6] proposed an object detection method that applying classification to detection as like Fully Convolutional Neural Network (FCN) [7] for satellite imagery.  They found the increase of  recall performance as the relative abundance of negative image decreases in the training data set. Our HRN model is equivalent to their model (Fig. 2), but the percentage of negative data is adjusted to achieve higher recall (Table. 1). In addition, we focus on the recall performance rather than precision during the training. HRN training process output a learned snapshot model each 10 epoch. From the many snapshot models generated by the HRN training process, we selected the model with the highest recall and with at least over 50\% precision. When selecting this model, we use validation dataset rather than training data set.

\begin{figure}[tb]
  \begin{center}
\vspace{-2cm}
  \includegraphics[bb=0 0 2000 929, scale=0.2]{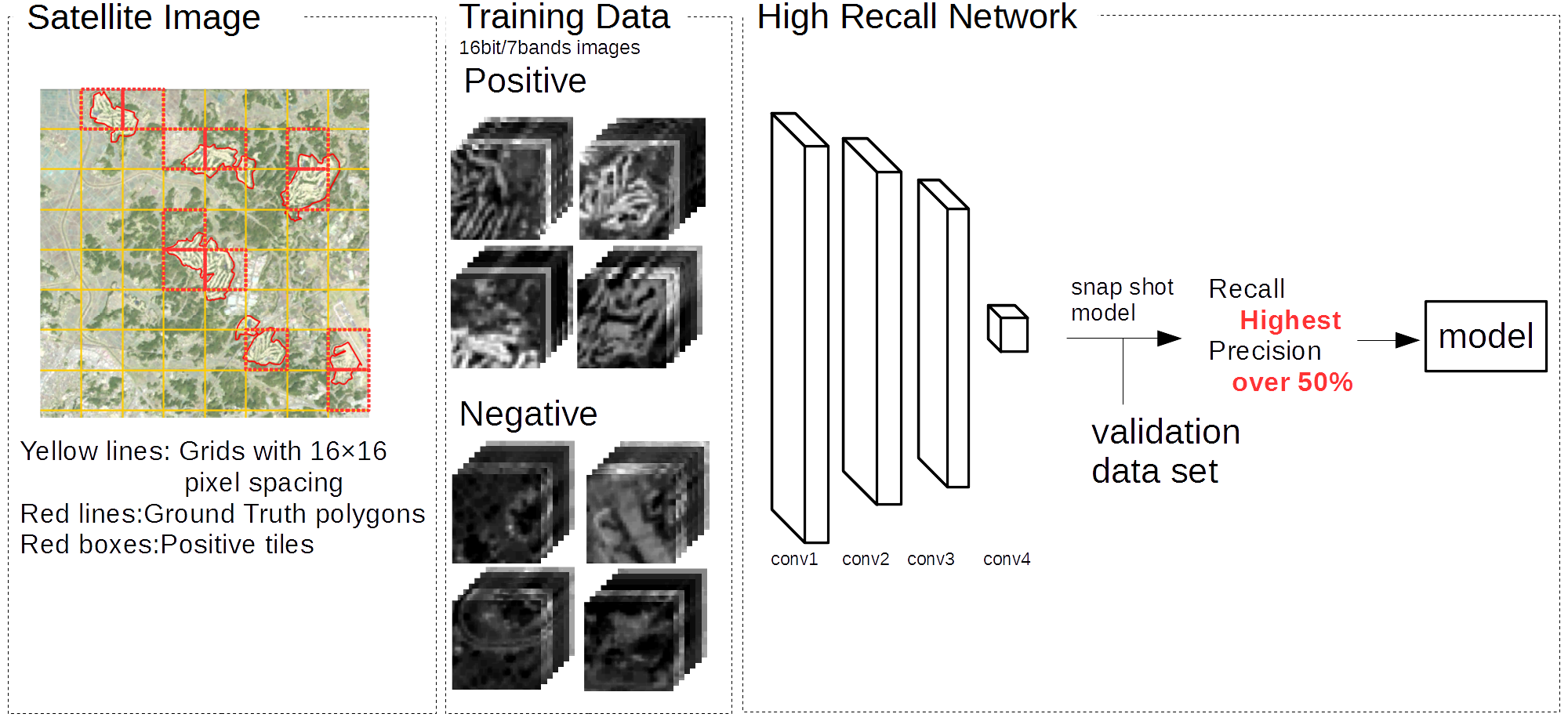}
  \caption{High Recall Network(HRN) has FCN structure consisting of 4 convolution layers. }
  \label{}
  \end{center}
\end{figure}

\subsection{High Precision Network (HPN) }
\label{ssec:subhead}
The HPN structure and training process is illustrated in Fig. 3. The HPN performs binary classification on the candidate regions resulted from HRN. Input data used in this process are derived from cropping HRN output into tiles with 64$\times$64 pixels (red rectangles in Fig. 3).

For HPN model training, positive images are generated by cropping satellite image of golf course region at the centroid from GT polygons. We conducted data augmentation by rotating each positive image in the step of 90$^\circ$ and then flipped in horizontally and vertically respectively. Negative images are produced by negative mining that crop a centroid of false positive regions generated by HRN. The number of convolution layers is increased to 8 to give the highest precision.

\begin{figure}[tb]
  \begin{center}
\vspace{-4.5cm}
  \includegraphics[bb=0 0 1754 1500, scale=0.23]{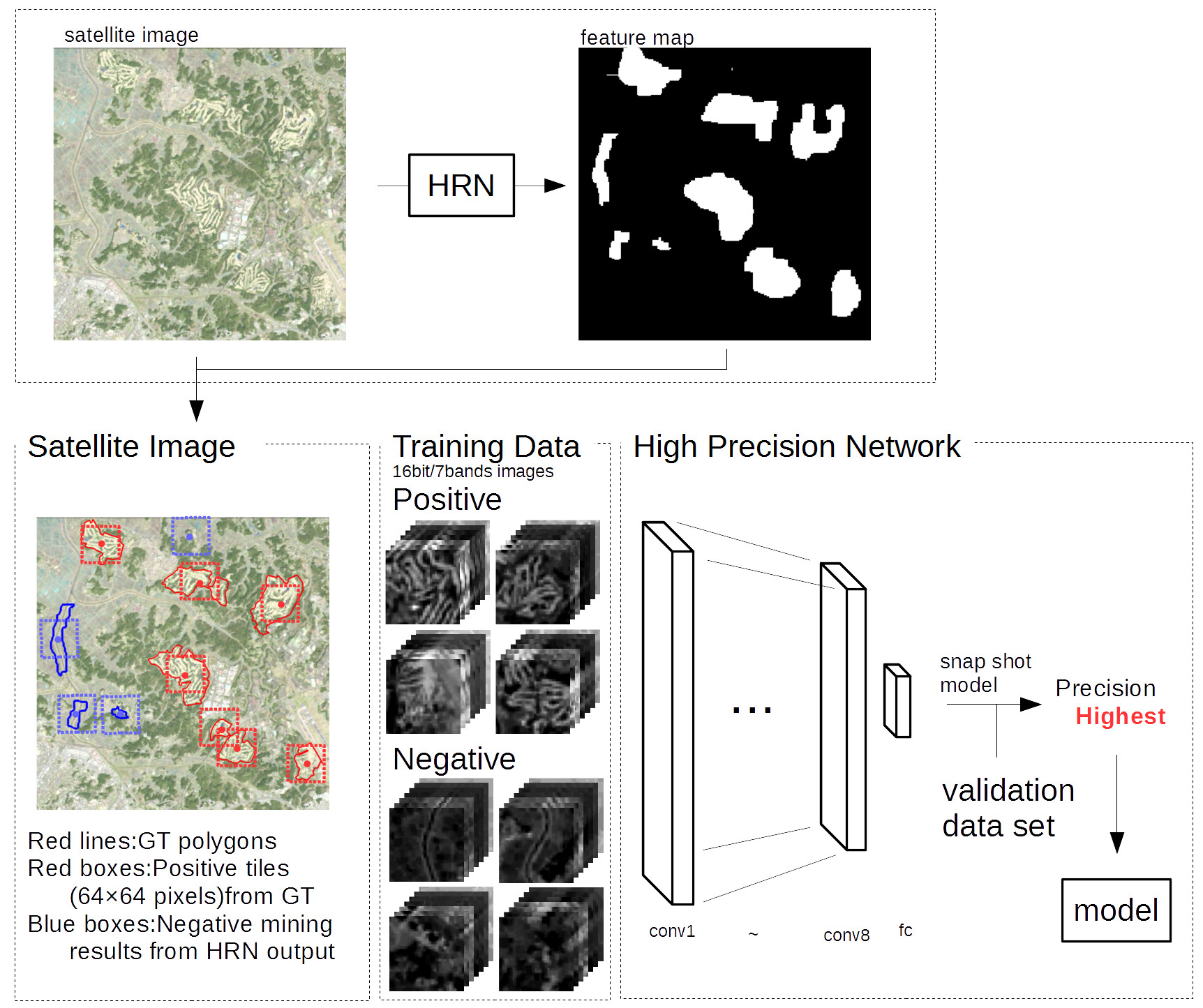}
  \caption{The High Precision Network (HPN)  comprises 8 convolutional layers and a fully connected layer. }
  \label{}
  \end{center}
\end{figure}

\section{EXPERIMENTS AND RESULTS}
\label{sec:pagestyle}
In this section, we conducted golf course detection experiments using our method as well as previous methods　(Ishii et al [6] and Uehara et al [1]). For direct comparison, we employed five scenes in Uehara et al [1] as testing data (Fig. 4).　It should be noted again that these testing data are not included in the training data. After removing cloudy areas, we classified the output tiles as true positives (TP) if it contains GT polygons and as false positive (FP) otherwise. It would be natural to regard GT polygons with no overlapping detected tiles as false negatives (FN). In Fig. 5, red lines are golf course regions from GT and red rectangles indicate the output tiles: (a) represents an example of TP in which a golf course was correctly detected; (b) represents FP in which an area was erroneously identified as a golf course; and (c) represents FN in which a golf course was not detected due to its atypical structure (a narrow course, no observable trees, etc.).

Table. 2 compares the performance of three methods measured by recall [ TP/(TP+FN) ], precision [TP/(TP+FP)] and F-measure [2$\times$precision$\times$recall/(precision+recall)]. When we use F-measure as a total performance proxy, our method showed 4\% more improvement than previous two methods. This improvement could be attributed to collaboration between HRN and HPN. In actual applications, F-measure may not be necessarily the best index to estimate the performance. Some applications would require an exhaustive list of the possible target candidates even with the lower precision, while others may need higher precision at the cost of lower recall.  Our framework could be easily adapted to diverse applications due to the explicit separation into two complementary networks.

\begin{table}[htb]
\begin{center}
\caption{Number of training data in experiments}
 \begin{tabular}{c||c|c|c|c} 
 \multicolumn{1}{}{} &
 \multicolumn{2}{c}{for HRN}
      & \multicolumn{2}{c}{for HPN} \\ 
     & positive & negative & positive & negative \\ \hline \hline
    training & 52276 & 936000 & 110314 & 298945 \\ \hline
    validation & 17620 & 4686120 & 35766 & 938271 \\ \hline
  \end{tabular}
  \label{dataset}
  \end{center}
\end{table}

\begin{figure*}[t]
  \begin{center}
\vspace{-3cm}
  \includegraphics[bb=0 0 1261 321, scale=0.68]{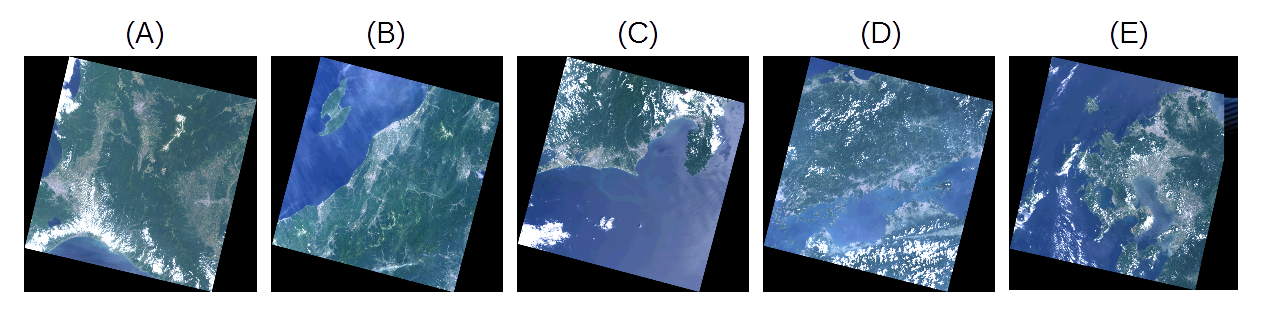}
  \caption{Test images used for the experiment. The scene IDs are, (A): LC81070302016189, (B): LC81080342015193, (C):LC81080362016196, (D): LC81110362015214, (E): LC81130372015212}
  \label{}
  \end{center}
\end{figure*}
\setlength\intextsep{0pt}

\begin{table}[htb]
\begin{center}
\caption{Experimental results}
  \begin{tabular}{c|c|c|c|c|c|c} 
     &A & B & C& D& E & total\\ \hline \hline
Recall \\ \hline
Ishii  &  0.333   & 0.736 & 0.743 & 0.903 & 0.889&0.759 \\ \hline
Uehara &  0.428   & 0.714 & 0.848 & 0.841 & 0.897&0.753 \\ \hline
    Our & 0.892 & 0.929& 0.833 & 0.937 & 0.908&0.901 \\ \hline \hline
Precision  \\  \hline 
Ishii   & 0.443 & 1.000& 1.000 & 0.981 & 0.986 &0.924\\ \hline
Uehara & 0.715 & 0.947& 0.924 & 0.910 & 0.911&0.895 \\ \hline
   Our & 0.456 & 0.956& 0.977 & 0.953 & 0.959&0.874 \\ \hline \hline
    F-measure   \\  \hline
    Ishii  & 0.380 & 0.848 & 0.852 & 0.940 & 0.935 &0.833\\ \hline
    Uehara & 0535 & 0.814 & 0.884 & 0.874 & 0.904 &0.818\\ \hline
       Our& 0.604 & 0.942 & 0.900 & 0.945 & 0.933&0.870 \\ \hline \hline
%    IOU   \\  \hline
%   Ishii et al. & 0.201 & 0.736 & 0.743 & 0.888 & 0.877&0.694 \\ \hline
%   Uehara et al& 0.366 & 0.687 & 0.792 & 0.776 & 0.825&0.692 \\ \hline
%    Our & 0.388 & 0.890 & 0.818 & 0.896 & 0.874&0.750 \\ \hline
  \end{tabular}
  \label{japan}
  \end{center}
\end{table}
%\vspace{-7mm}

\begin{figure}[!h]
  \begin{center}
\vspace{-5cm}
  \includegraphics[bb=0 0 1193 1147, scale=0.33]{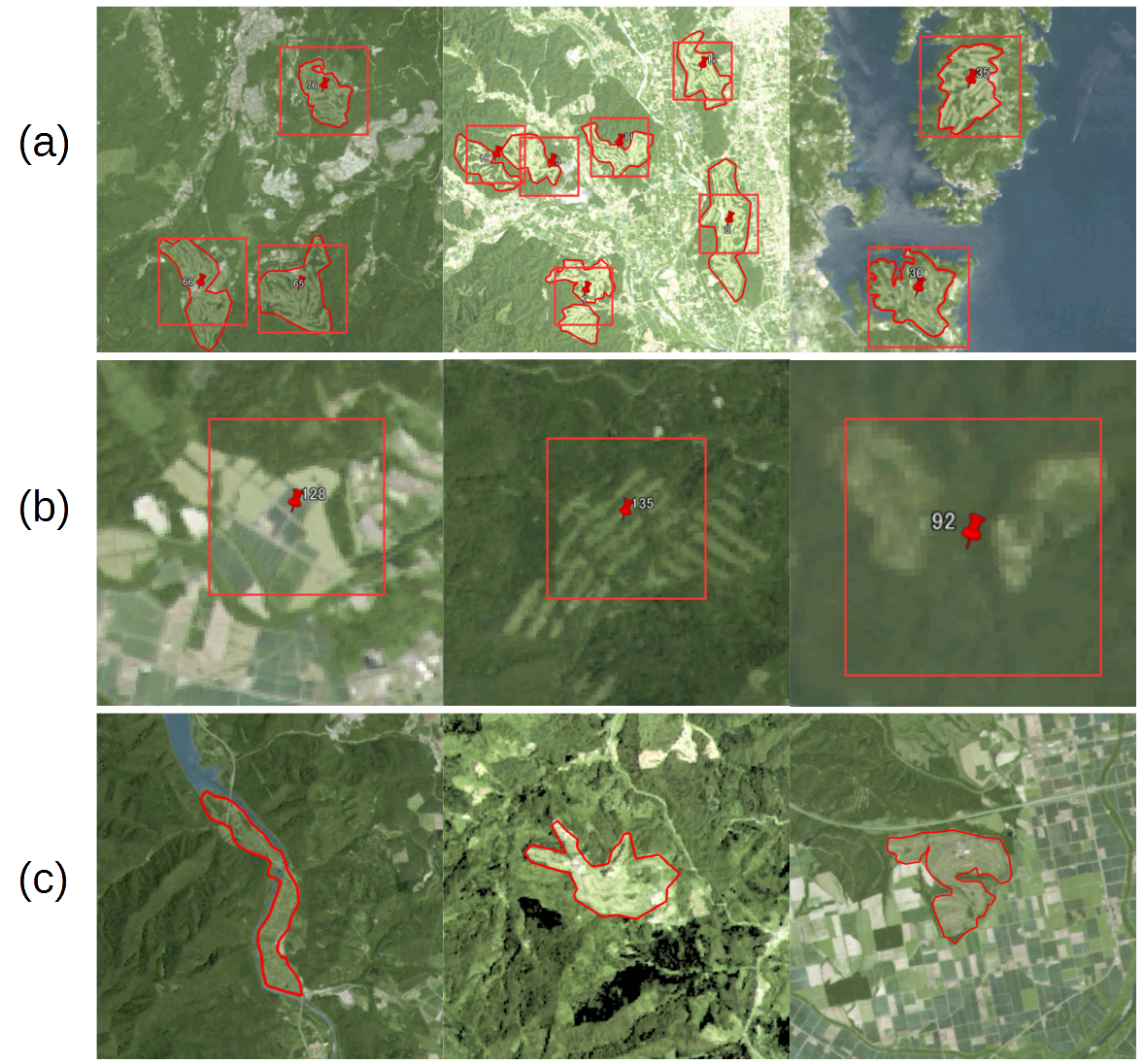}
  \caption{An example of experimental results with the proposed method: (a) True Positives (b) False Positives (c) False Negatives. The red lines represent GT polygons of the golf course, while rectangles indicate the output tiles (64x64 pixels) with detection.}
  \label{}

  \end{center}
\end{figure}
\setlength\intextsep{0pt}
\section{CONCLUSION}
\label{sec:typestyle}
In this paper, we proposed a method to detect arbitrary objects in satellite imagery . The method entailed an integration of two convolutional neural networks (CNN) devoted to high recall and high precision, respectively. We customized each CNN through network structure, selection of the input training data and target parameters for optimization. Our method showed a 4\% overall improvement compared to previous methods. Furthermore, we can flexibly tune the relative importance of recall and precision by balancing two networks.

In the future work, we will
(1) expand the target areas to a global level and evaluate the generalization performance. 
(2) increase the tile size for HPN to apply the state-of-the-art CNN for general images [10][11].
(3) make a comparison with the state-of-the-art two step algorithm for general images, such as Faster R-CNN [8].

\section{ACKNOWLEDGMENT}
\label{sec:majhead}
This paper is based on results obtained from a project commissioned by the New Energy and Industrial Technology Development Organization (NEDO).

%\section{REFERENCES}
%\label{sec:ref}

% BiBTeX files (here: strings, refs, manuals). The IEEEbib. bst bibliography
\bibliographystyle{IEEEbib}
\bibliography{strings, refs}

\end{document}